\documentclass[sigconf, screen]{acmart}

\AtBeginDocument{%
  }

\setcopyright{acmlicensed}
\copyrightyear{2025}
\acmYear{2025}
\setcopyright{cc}
\setcctype{by}
\acmConference[MM '25]{Proceedings of the 33rd ACM International Conference
on Multimedia}{October 27--31, 2025}{Dublin, Ireland}\acmBooktitle{Proceedings of the 33rd ACM International Conference on
Multimedia (MM '25), October 27--31, 2025, Dublin,
Ireland}\acmDOI{10.1145/3746027.3758174}
\acmISBN{979-8-4007-2035-2/2025/10}

\usepackage{tikz}
\usepackage{graphicx}
\usepackage{wrapfig}
\usepackage{booktabs}
\usepackage{adjustbox}
\usepackage{multirow}
\usepackage{xspace}
\usepackage{array}
\usepackage{bm}
\usepackage{bbm}
\usepackage{amsthm,amsmath}
\usepackage{cleveref}
\usepackage{algorithm}
\usepackage{algorithmic}

\usepackage{listings}
\usepackage{wrapfig}
\usepackage{subcaption}
\usepackage{url}
\usepackage{colortbl}
\usepackage{adjustbox}
\usepackage{makecell}
\usepackage{pifont}
\usepackage{tcolorbox}
\usepackage{setspace}
\usepackage{fancybox}
\usepackage{tocloft}
\usepackage{etoc}
\usepackage{enumitem}
\usepackage{enumerate}
\usepackage{tablefootnote}

\definecolor{mymauve}{rgb}{0,0.6,0}
\begin{document}

\title{Towards Modality Generalization: A Benchmark and Prospective Analysis}

\author{Xiaohao Liu}
\affiliation{%
  \institution{National University of Singapore}
  \city{Singapore}
  \country{Singapore}
}
\email{xiaohao.liu@u.nus.edu}

\author{Xiaobo Xia}
\affiliation{%
  \institution{National University of Singapore}
  \city{Singapore}
  \country{Singapore}
}
\email{xbx@nus.edu.sg}
\authornote{Corresponding author.}

\author{Zhuo Huang}
\affiliation{%
  \institution{The University of Sydney}
  \city{Sydney}
  \country{Australia}
}
\email{zhuohuang.ai@gmail.com}

\author{See-Kiong Ng}
\affiliation{%
  \institution{National University of Singapore}
  \city{Singapore}
  \country{Singapore}
}
\email{seekiong@nus.edu.sg}

\author{Tat-Seng Chua}
\affiliation{%
  \institution{National University of Singapore}
  \city{Singapore}
  \country{Singapore}
}
\email{dcscts@nus.edu.sg}

\renewcommand{\shortauthors}{Liu et al.}

\begin{abstract}
Multi-modal learning has achieved remarkable success by integrating information from various modalities, achieving superior performance in tasks like recognition and retrieval compared to uni-modal approaches. However, real-world scenarios often present novel modalities that are unseen during training due to resource and privacy constraints, a challenge current methods struggle to address. This paper introduces Modality Generalization (MG), which focuses on enabling models to generalize to unseen modalities. We define two cases: Weak MG, where both seen and unseen modalities can be mapped into a joint embedding space via existing perceptors, and Strong MG, where no such mappings exist. To facilitate progress, we propose a comprehensive benchmark featuring multi-modal algorithms and adapt existing methods that focus on generalization. Extensive experiments highlight the complexity of MG, exposing the limitations of existing methods and identifying key directions for future research. Our work provides a foundation for advancing robust and adaptable multi-modal models, enabling them to handle unseen modalities in realistic scenarios.

\end{abstract}



\received{20 February 2007}
\received[revised]{12 March 2009}
\received[accepted]{5 June 2009}

\maketitle

\section{Introduction}
The world is perceived and interpreted through diverse modalities, \textit{e.g.}, images, videos, text, and audio~\cite{wu2024deep,zhang2023on,lu2024computational,lu2023theory,wu2024next,luo2024mmevol,luo2024deem,zhou2025towards,zhou2024few,zhou2025dreamdpo}. Each modality offers a unique perspective, which captures patterns and describes objects from certain physical viewpoints. Against this backdrop, multi-modal learning has emerged as a powerful approach to developing models capable of processing and integrating information across multiple modalities~\cite{huang2021makes,xu2023multimodal,ektefaie2023multimodal,bayoudh2022survey,liang2024foundations,luo2025vcm}. Compared to traditional single-modal methods, multi-modal learning not only achieves superior performance in conventional tasks~(\textit{e.g.}, object recognition~\cite{radford2021learning} and detection~\cite{chen2022multimodal}), but also opens up opportunities for exploring novel tasks enabled by cross-modal interactions~(\textit{e.g.}, cross-modal retrieval~\cite{zhen2019deep}, audio-visual learning~\cite{zhu2021deep}, and visual question answering~\cite{cadene2019murel}).

\begin{figure}[!t]
    \centering
    \includegraphics[width=1.0\linewidth]{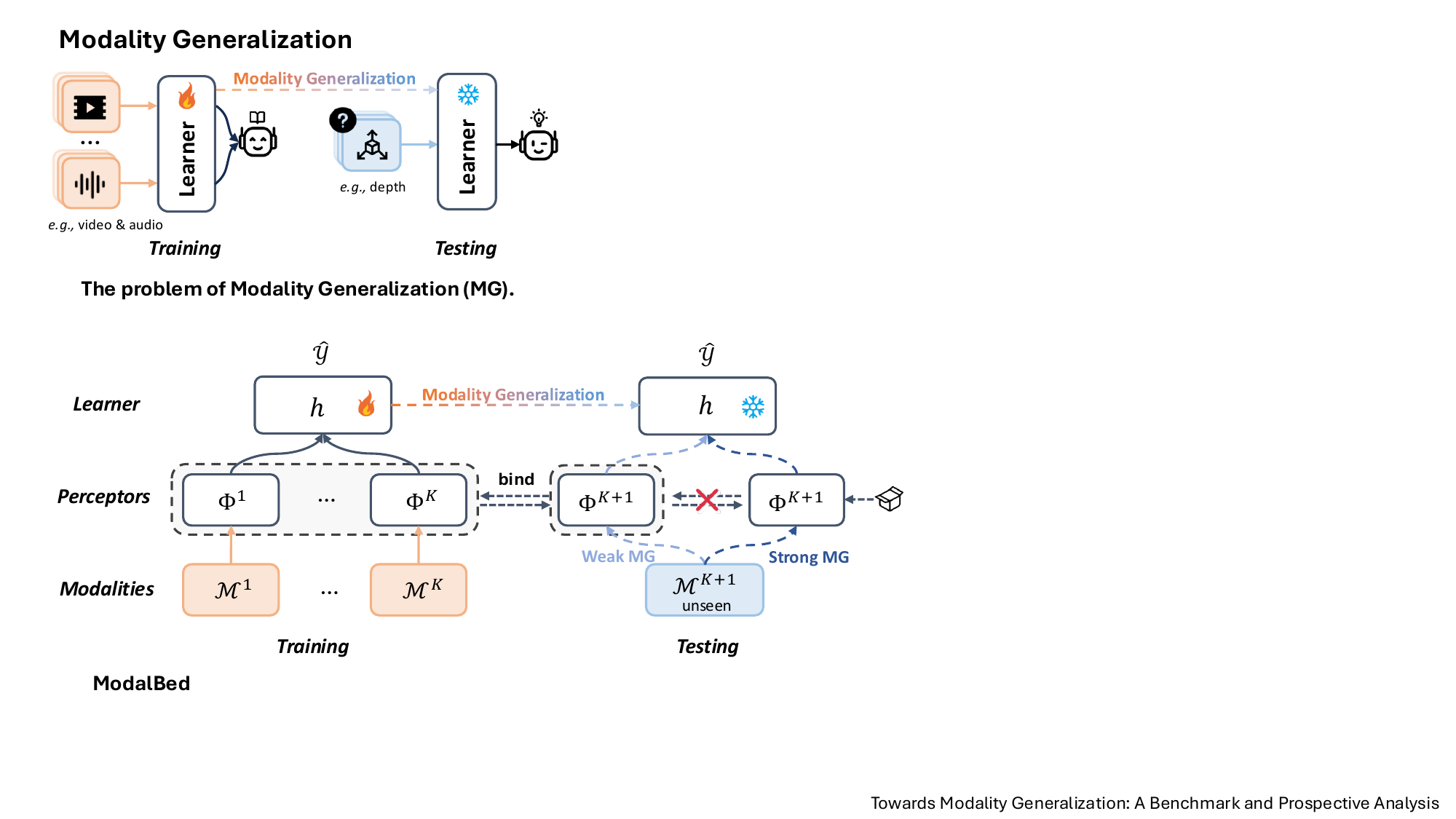} 
    \caption{\textbf{The problem of Modality Generalization (MG).} The learner trained on multiple modalities (\textit{e.g.}, video or audio) is capable of performing well on unseen modalities (\textit{e.g.}, depth) during testing. }
    \label{fig:mg_problem}
    \vspace{-4mm}
\end{figure}

\begin{table*}[!t]
    \centering
    \caption{A comparison of different multi-modal learning and generalization problems. Here ``Correspondence'' indicates instance-level modal correspondence, \textit{i.e.}, multi-modal data are paired with the same instance. We use ``\ding{51}'' to denote the correspondence is needed or employ ``\ding{55}'' otherwise. }
    \renewcommand{\arraystretch}{1}
    \setlength{\tabcolsep}{2pt}
    \resizebox{0.75\textwidth}{!}{
    \begin{tabular}{c|c|ccc}
    \toprule
    \textbf{Problem}     &  \textbf{Reference} & \textbf{Train Inputs} & \textbf{Test Inputs} & \textbf{Correspondence} \\\midrule
    Cross-Modal Fine-Tuning & \cite{cai2024enhancing, shen2023cross, dong2023simmmdg} & Known & Known \& Seen & \ding{51} \\
    Cross-Modal Generalization & \cite{liang2021cross, xia2024achieving} & Known & Unknown \& Seen & \ding{51} \\
    MML w/o labeled Multi-Modal Data & \cite{liang2023multimodal} & Known & Unknown \& Seen & \ding{51}\\
    Modality Connection & \cite{wang2023connecting, wang2023extending, ma2019unpaired} & Known & Known \& Seen & \ding{55} \\
    Out-of-Modal Generalization & \cite{anonymous2024towards} & Known & Unknown \& Seen & \ding{55}\\ \midrule
    \multirow{2}{*}{Modality Generalization} & Weak MG & Known & Known \& Unseen & \ding{55} \\
    & Strong MG & Known & Unknown \& Unseen & \ding{55}
    \\ \bottomrule
    \end{tabular}
    }
    \label{tab:problem_setups}
\end{table*}

In real-world scenarios, it is often difficult or even infeasible to include all possible modalities in the training process due to limitations in resources, policies, hardware, or processing techniques. For instance, acquiring and processing certain modalities such as tactile data, thermal imaging, or privacy-sensitive data like medical records often requires specialized sensors, complex setups, and strict compliance with privacy regulations, making their inclusion in training problematic. However, models frequently encounter data from modalities that were not part of the training set. This poses a significant challenge for current multi-modal learning methods, as they often rely on well-aligned and pre-defined modalities during training. Without proper generalization strategies, the inability to handle such unseen modalities will limit the robustness and applicability of multi-modal learning.

Aware of the mentioned situation and issue, we highlight the problem of modality generalization~(MG) in this paper, which focuses on enabling models to adapt to and perform well on unseen modalities beyond those available during training~(see Figure~\ref{fig:mg_problem}). Besides, we consider two practical cases of the problem: Weak MG and strong MG. Specifically, Weak MG refers to the case where some modalities are unseen during training but can be inferred by existing perceptors linked to training modalities. 
Such perceptors can be realized by many existing modality binding models that are pre-trained with paired modality data to map various modalities (\textit{i.e.}, training and testing modalities) into a joint embedding space\footnote{Note that sharing a joint embedding space across different modalities does not inherently address the MG problem. Further explanations are provided in Section~\ref{sec:limitation-joint-embedding}.}. By taking advantage of these perceptors, the feature embeddings of unseen modalities can be directly extracted which enables subsequent adaptation and generalization for solving MG. As a comparison, strong MG describes the case where some modalities are neither seen during training nor perceptible by existing models. For example, a model trained on the text and audio data may encounter a new modality like bio-signals, with no perceptors to map it into the joint embedding space. This requires the model to generalize across modalities without direct alignment or shared representations, demanding a higher level of robustness.

To draw attention to this critical problem and facilitate its exploration, we have thoroughly analyzed the concept of modality generalization (MG), highlighting its distinctions and connections with prior work. To enable reproducible research and fair comparisons, we introduce a comprehensive benchmark tailored for MG. This benchmark includes a curated set of multi-modal learning algorithms as well as methods adapted from other domains that are capable of addressing MG challenges. Through extensive experiments using the benchmark, we provide detailed discussions of the results, illustrating the complexity of the MG problem. Our findings highlight the limitations of existing approaches in handling unseen modalities and underscore the need for novel strategies to improve robustness and adaptability. Additionally, we outline key points for future research, offering insights into potential directions for advancing the field of modality generalization. We hope this work serves as a valuable resource for the research community and accelerates progress in this challenging and impactful area.

\begin{figure*}[t]
    \centering
    \includegraphics[width=0.72\linewidth]{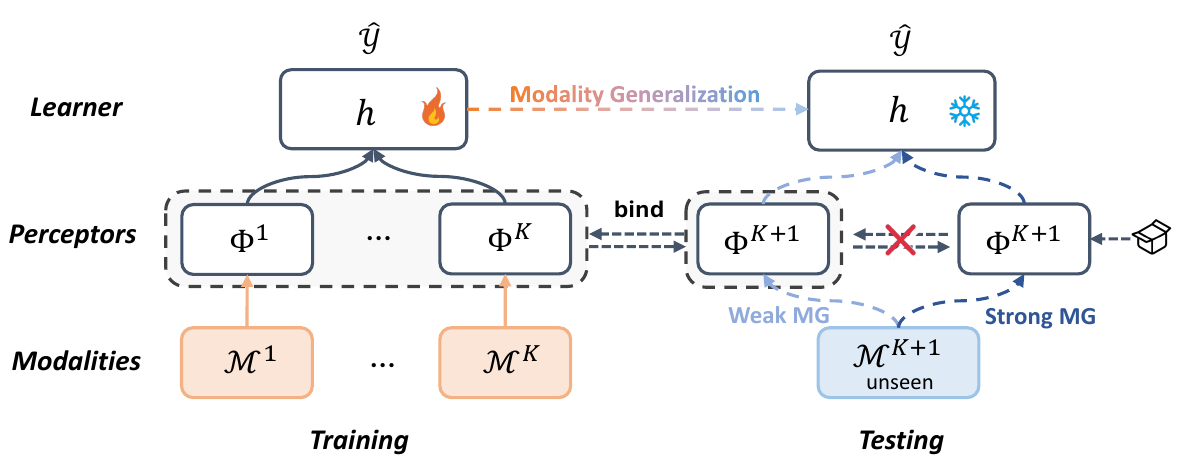}
    \vspace{-5mm}
    \caption{The framework of \textsc{ModalBed}, involving two cases of Weak MG and Strong MG. These two cases share a common training process, where perceptors $\{\Phi^1,\dots, \Phi^K\}$ extract training modalities $\{\mathcal{M}^1,\dots, \mathcal{M}^K\}$, and a learner $h$ is trained for follow-up tasks. Notably, in Weak MG, the testing perceptor $\Phi^{K+1}$ is pre-trained with the same objective as the training perceptors to generate joint embeddings. As a comparison, in Strong MG, the testing perceptor is kept \textit{isolated} from training perceptors.}
    \vspace{-3mm}
    \label{fig:weak_strong_MG}
\end{figure*}

\section{Related Work}

\subsection{Modality Binding} 
The modality binding strategy aims to learn a joint embedding space across different modalities. CLIP~\cite{radford2021learning} pioneers the alignment of image and language data and makes them share an embedding space for downstream tasks, \textit{e.g.}, zero-shot classification~\cite{qian2024online} and open-vocabulary segmentation~\cite{wang2024open}, and inspires lots of subsequent works on modality alignment~\cite{li2022blip,li2023blip,song2023bridge,guo2023point,yang2024binding,zhou2024aligning,zhang2023frame,liu2025continual, liu2025principled}. Recently, ImageBind~\cite{girdhar2023imagebind} proposes to utilize vision modality as a bridge to unify various modalities into a shared embedding space. LanguageBind~\cite{zhu2024languagebind} employs language as an alternative solution to bridge different modalities similarly. Moreover, UniBind~\cite{lyu2024unibind} makes the alignment centers modality agnostic and further learns a unified and balanced embedding space, empowered by large language models. Note that modality binding can provide perceptors to some modalities and benefits in solving the MG problem. 

\subsection{Multi-Modal Learning and Generalization} Previous works have focused on the objective of leveraging the knowledge from some modalities and generalizing it to another one. For instance, cross-modal fine-tuning~\cite{cai2024enhancing, shen2023cross, dong2023simmmdg} operates similarly to transfer learning by aligning the training modality data distribution with the testing modality data, utilizing a shared model architecture. Cross-modal generalization~\cite{liang2021cross, xia2024achieving} employs distinct encoders and aims to generalize across different modality data derived from the same instance. Besides, multi-modal learning (MML) w/o labeled multi-modal data~\cite{liang2023multimodal} proposes a setting with unpaired labeled training modality data and testing modality data, supplemented by unlabeled paired multi-modal data for learning modality interactions. Modality connection~\cite{wang2023connecting, wang2023extending, ma2019unpaired} aims to solve the mismatched modality across datasets, but it does not explore unseen modalities. Recent out-of-modal generalization studies how to adapt to an unknown modality without given instance-level modal correspondence or scarcely paired correspondence in some cases. However, the setting of our modality generalization is different from theirs, as they require incorporating previously unseen modality data into the training process, whereas our setting does not rely on training with the unseen modality. To further distinguish the problem settings between ours and existing works, we denote the modal knowledge possessed by pre-trained perceptors as ``known'', and the modalities shown in the training set as ``seen'', otherwise denoted as ``unknown'' and ``unseen'', respectively, in Table~\ref{tab:problem_setups}.

\subsection{Domain Generalization} The process of training a model on multiple source domains to enable generalization to unseen target domains is known as domain generalization~\cite{wang2022generalizing}. Compared to domain adaption~\cite{pan2010domain}, domain generalization is more challenging because target domain data cannot be accessed during training. Previous studies have handled domain generalization from a series of perspectives~\cite{wang2022generalizing,dong2023simmmdg}, \textit{e.g.}, data manipulation~\cite{tobin2017domain,zhang2017mixup,zhou2020deep}, learning domain-invariant representations~\cite{ganin2016domain,li2018deep,lin2022zin}, using gradient operation~\cite{huang2020self}, with meta-learning~\cite{li2018learning}, and \textit{etc}. Readers
can refer~\cite{wang2022generalizing} for more details on domain generalization. Apparently, domain generalization and modality generalization share similar objectives in enabling models to generalize to unseen settings. Many methodologies developed for domain generalization, such as invariant learning and adversarial alignment, can inspire solutions for modality generalization. In addition, the benchmark in domain generalization~(\textit{e.g.}, \textsc{DomainBed}~\cite{gulrajani2021search}) can help modality generalization implementation. 

\section{Preliminaries}
\subsection{Problem Formulation}
In the problem of modality generalization~(MG), during training, we are given a set of known modalities $\{\mathcal{M}^1,\ldots,\mathcal{M}^K\}$ where $\mathcal{M}^{k\in\{1,\ldots,K\}}=\{(\mathbf{x}^k_i,y^k_i)\}_{i=1}^{N_k}$ includes a total of $N_k$ labeled training modality examples. Afterward, in testing, we have an unseen modality $\mathcal{M}^{K+1}=\{(\mathbf{x}_i^{K+1})\}_{i=1}^{N_{K+1}}$ that contains $N_{K+1}$ unlabeled testing instances. MG aims to learn a predictor only with training modalities, which can perform well at the testing modality.  In this paper, we mainly focus on the classification task, where only the testing modality is involved for prediction. The cases of Weak and Strong MG are detailed below. 

\subsection{Weak MG}
In this case, we have some available perceptors for training and testing modalities, which can map them into the joint embedding space for downstream tasks. We denote the perceptors about training modalities as $\{\Phi^1,\ldots,\Phi^K\}$, where $\Phi_i$ corresponds to the $i$-th modality. The perceptor about the unseen testing modality is denoted by $\Phi^{K+1}$. Many existing modality-binding models can realize these perceptors, \textit{e.g.},  ImageBind and LanguageBind, which are trained with paired modality data (\textit{e.g.}, image-centric or language-centric) to exploit the joint embedding space across modalities. Afterward, the embeddings of the training and testing modality data can be obtained by $\mathbf{z}^k_i=\Phi^k(\mathbf{x}^k_i)$ and $\mathbf{z}^{K+1}_{i}=\Phi^{K+1}(\mathbf{x}^{K+1}_{i})$. 
Note that we do not target the improvement of the perceptors and therefore freeze their parameters for embedding extraction.  On top of the above models, we define a learner $h$ trained on the embeddings of the training modality data. This learner is then used to generalize to the unseen testing modality for follow-up tasks.

\subsection{Strong MG} 
In this situation, we still access the perceptors of training modalities, \textit{i.e.}, $\{\Phi^1,\ldots,\Phi^K\}$ for embedding extraction. Differently, despite various modalities involved with paired data in modality-bind perceptors pre-training, novel modalities emerge with few or even no paired data, exhibiting a more challenging scenario for modality generalization. In this case, we do not have a perceptor for the testing modality to map it into the joint embedding space of training modalities. Therefore, we have to employ some perceptors solely pre-trained within the testing modality data $\mathcal{M}^{K+1}$ for its embedding extraction and finish follow-up tasks with the trained learner together. Here, embedding spaces differ between training and testing modalities, and we consider that Strong MG approaches have the potential to uncover the hidden \emph{invariant features} while bypassing the embedding space variants. 
The illustrations about these two cases are shown in Figure~\ref{fig:weak_strong_MG}.

\noindent\textbf{$\bullet$ Difference from Weak MG. }
In Strong MG, the perceptor for the testing modality is trained via self-supervised learning, which provides information relevant to the unseen modality. Therefore, the challenge lies in how the algorithm captures invariances between the training modalities, thus generalizing to the unseen modality. This is more difficult than Weak MG, which builds upon existing modality binding methods.

\noindent\textbf{$\bullet$ Potentials of Strong MG.} Note that Strong MG is very practical in real-world scenarios where modalities emerge as technology advances. This further inspires us to incorporate Strong MG into our benchmark. In fact, previous works such as~\cite{tjandrasuwita2025understanding, huh2024position} also support the potential of Strong MG, where independently trained perceptors of increasing scale can be implicitly aligned with each other.

\begin{table*}[t]
\centering
\caption{The important statistics of experimental datasets. We use ``\ding{51}'' for observable modality in a dataset and ``-'' otherwise. }
\vspace{-2mm}
\label{tab:dataset}
\resizebox{0.86\textwidth}{!}{
\begin{tabular}{l|ccccc|c|c} 
\toprule
\multirow{2}{*}{\textbf{Dataset}} & \multicolumn{5}{c|}{\textbf{Modalities}}                                   & \multirow{2}{*}{\textbf{\#. Instance}} & \multirow{2}{*}{\textbf{\#. Classes}}  \\
& \textbf{Video (Vid)} & \textbf{Audio (Aud)} & \textbf{Depth (Dep)} & \textbf{Image (RGB)} & \textbf{Language (Lan)} &                                        &                                        \\ 
\midrule
\textbf{MSR-VTT}& \ding{51} &\ding{51} & - & - &\ding{51} & 10,000 & 20 \\ 
\midrule
\textbf{NYUDv2}& - & - &\ding{51} &\ding{51} &\ding{51} & 47,584 & 23 \\ 
\midrule
\textbf{VGGSound-S} &\ding{51} &\ding{51} & - & - &\ding{51} & 10,000 & 310 \\
\bottomrule
\end{tabular}
}
\end{table*}

\subsection{Limitations of Joint Embedding Spaces}
\label{sec:limitation-joint-embedding}
Note that sharing a joint embedding space across different modalities does not inherently solve the MG problem, as it often fails to effectively learn \textit{invariant} features across modalities for generalization. Invariant features, which capture shared semantic or structural information across different modalities, are essential for generalization. However, it is hard to learn only with joint embedding spaces. For example, if the alignment model lacks sufficient capacity or if the paired data between modalities is sparse, the alignment may focus on local correlations specific to individual modalities rather than extracting globally shared features. Besides, the inherent heterogeneity between modalities poses significant challenges. For example, the feature space of language data, which emphasizes semantic relationships, differs greatly from that of EEG signals, which encode neural activity patterns. Such heterogeneity makes it difficult for the alignment model to map these modalities into a joint embedding space while preserving their shared invariant features. Without successfully learning these invariant features during training, the shared embedding space cannot generalize well to unseen modalities during testing. This limitation underscores the core challenge of the MG problem, highlighting the need for strategies that explicitly address the learning and transfer of invariant features across diverse and heterogeneous modalities.

\vspace{-0.5mm}
\section{\textsc{ModalBed}: A PyTorch Tester for Modality Generalization}

At the heart of our large-scale experiments is \textsc{ModalBed}, a PyTorch-based framework designed to facilitate reproducible and solid research in modality generalization at:

{\centering 
\href{https://github.com/Xiaohao-Liu/ModalBed}{\textsc{ModalBed} Repository Link}
\par}

The initial release mainly includes three multi-modal learning/generalization algorithms and 10 domain generalization algorithms (built upon \textsc{DomainBed}). \textsc{ModalBed} is an ongoing project that will be continually updated with new results, algorithms, and datasets. Contributions from fellow researchers through pull requests are highly encouraged and welcomed.

\noindent\textbf{$\bullet$ Continuing \textsc{ModalBed}.}
\label{sec:continual_modalbed}
\textsc{ModalBed} is believed to be a valuable tool for benchmarking modality generalization algorithms. Its continued development is essential for advancing the research field. Future updates to \textsc{ModalBed} include: (1) \textit{integration of new datasets and modalities}: regularly update the platform with datasets involving novel modalities or their combinations to reflect realistic challenges; (2) \textit{implementation of cutting-edge algorithms}: expand the repository to include emerging techniques in multi-modal and domain/modality generalization research; (3) \textit{collaborative contributions}:  foster an open-source community to encourage researchers to contribute algorithms, datasets, and benchmarks, ensuring \textsc{ModalBed} remains comprehensive and up-to-date.

\section{Experimental Setups}

\subsection{Datasets} 
We consider datasets with at least three modalities: (1) \textbf{MSR-VTT}~\cite{xu2016msr}, which includes videos and text descriptions. Here we break down the videos into video frames and the audio data following~\cite{akbari2021vatt, lyu2024unibind}. (2) \textbf{NYUDv2}~\cite{silberman2012indoor}, which includes RGB images, depth data, and class names that can be transformed into language descriptions as well. (3) \textbf{VGGSound}~\cite{chen2020vggsound}, which contains video data, corresponding sound, and language descriptions. Note that we randomly sample instances from VGGSound to construct a smaller-scale dataset, enabling more efficient benchmarking. We name this smaller-scale set as VGGSound-S. The important statistics of these used datasets are shown in Table~\ref{tab:dataset}. More datasets will be included in future updates.

For each dataset, every modality can serve as the unseen/testing modality. Therefore, we construct different subdatasets based on the number of modalities. For example, MSR-VTT has three variants with video, audio, and language as corresponding testing modalities. 
Therefore, we have to design 6 variants (3 modalities for each dataset 
$\times$ 2 MG settings) for each dataset for a comprehensive benchmark.
All these subdatasets are used to train the model separately to avoid information leakage.

\subsection{Perceptors} 
We utilize three types of perceptors to process different modalities: (1) \textbf{ImageBind}~\cite{girdhar2023imagebind}; (2) \textbf{LanguageBind}~\cite{zhu2024languagebind}; (3) \textbf{UniBind}~\cite{lyu2024unibind}. Generally speaking, with paired data, these perceptors are obtained by learning a unified representation space across different modalities. Specifically, ImageBind leverages the binding property of images and learns a unified shared representation space by utilizing multiple types of image-paired data. By aligning each modality’s embedding to image embeddings, ImageBind achieves emergent alignment across diverse modalities without requiring all modalities to co-occur. Afterward, LanguageBind takes the language as the bind across different modalities and aligns the embeddings of the other modalities to corresponding language embeddings, implementing multi-modal semantic alignment. Furthermore, UniBind does not treat the image or language as the central modality. Differently, it makes the alignment centers modality agnostic and further learns a unified and balanced representation space. 
For weak modality generalization, the perceptors for both training and testing are available and sourced from the modality bind method. For strong modality generalization, the perceptors for training modalities still come from the modality bind. The perceptor of the testing modality is obtained by self-supervised learning only with the data of this modality. 
In this work, we use contrastive learning~\cite{chen2020simpclr} to implement self-supervised learning. 
Specifically, we mask parts of the loaded data. Text data is represented as token indices, and other modalities are loaded as matrices, with a masking ratio of 0.3 to generate two views for each instance. Masked elements are set to zero. We then apply contrastive learning to pull these two views closer in the representation level.

\subsection{Algorithms} 
Two types of algorithms are considered for the modality generalization task. 
The first type refers to algorithms specifically tailored for multi-modal learning~(\textbf{MML}), aiming to integrate information from diverse modalities to accomplish specific tasks effectively. 
Here we include algorithms like feature concatenation (\textbf{Concat}), On-the-fly Gradient Modulation (\textbf{OGM}~\cite{peng2022balanced}), and Dynamically Learning Modality Gap (\textbf{DLMG}~\cite{yangfacilitating}). 
The second type pertains to the algorithms of domain generalization~(\textbf{DG}), which aim to learn how to aggregate information across different domains, enabling generalization across known and unseen domains. 
We treat different modalities as different domains and then apply domain generalization algorithms. 
We initially include 10 algorithms: 
Empirical Risk Minimization (\textbf{ERM}~\cite{vapnik1998statistical}), Invariant Risk Minimization (\textbf{IRM}~\cite{arjovsky2019invariant}), Inter-domain Mixup (\textbf{Mixup}~\cite{yan2020improve}), Class-conditional DANN (\textbf{CDANN}~\cite{li2018deep}), Style Agnostic Networks (\textbf{SagNet}~\cite{nam2021reducing}), Information Bottleneck (\textbf{IB$_\text{ERM}$}~\cite{ahuja2021invariance}), Conditional Contrastive Adversarial Domain (\textbf{CondCAD}~\cite{ruan2021optimal}), Empirical Quantile Risk Minimization (\textbf{EQRM}~\cite{eastwood2022eqrm}),
Improved Empirical Risk Minimization (\textbf{ERM++}~\cite{teterwak2023erm++}),
and 
Uniform Risk Minimization (\textbf{URM}~\cite{krishnamachariuniformly}).

\subsection{Model Selection Methods} 
Differences in results due to inconsistent tuning practices can be mistakenly attributed to the algorithms under evaluation, complicating fair comparisons and masking true performance differences. This issue becomes even more pronounced in modality generalization, where the lack of a validation set that is identically distributed to the testing modality data makes model selection, such as choosing hyperparameters, training checkpoints, or architecture variants, particularly challenging. Without proper tuning practices and robust validation mechanisms, fair assessments and progress in modality generalization remain difficult to achieve. To benchmark modality generalization comprehensively and fairly, we choose a series of model selection methods, including \textbf{training-modality validation (TM)}, \textbf{leave-one-modality-out cross-validation (LOO)}, and \textbf{testing-modality validation (Oracle)}. 

\begin{figure*}
    \centering
    \includegraphics[width=\linewidth]{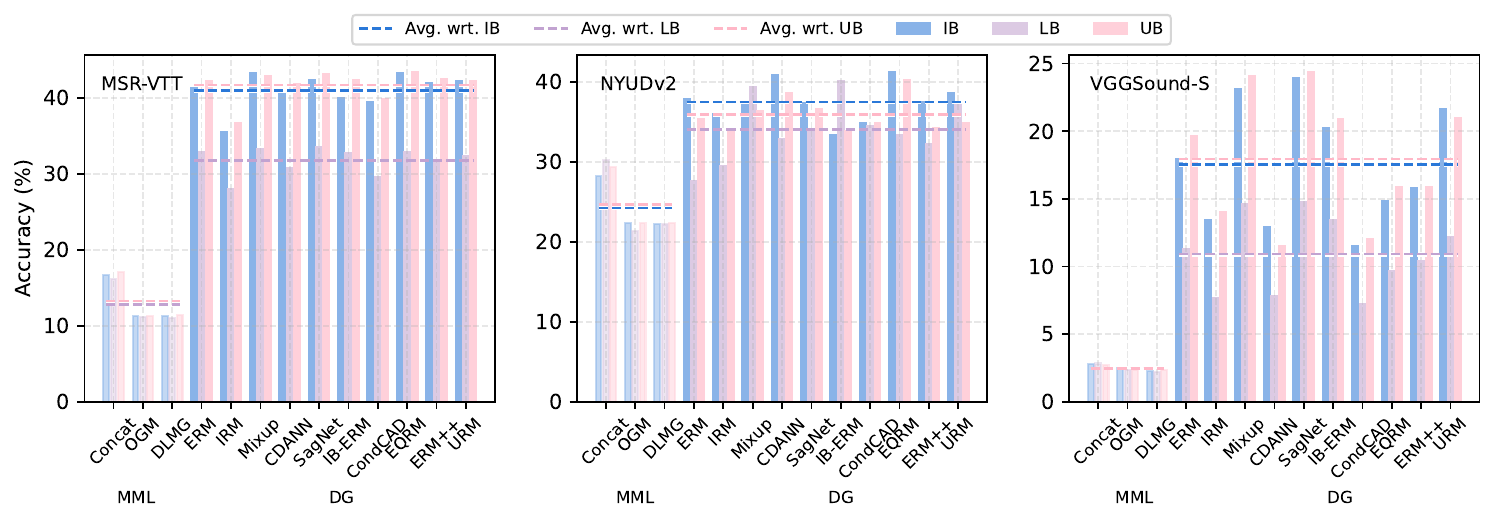}
    \vspace{-8mm}
    \caption{Weak MG performance comparison for diverse algorithms with three perceptors (``IB''$\leftrightarrow$ ImageBind, ``LB''$\leftrightarrow$ LanguageBind, and ``UB''$\leftrightarrow$ UniBind). 
    The dashed line indicates the averaged performance for two categories of algorithms (MML and DG) \textit{w.r.t.} different perceptors. 
    The bar represents the average performance of each algorithm across different modalities.}
    \vspace{-3mm}
    \label{fig:weakMG_bar}
\end{figure*}

\begin{figure*}[t]
    \centering
    \begin{minipage}[t]{0.34\linewidth}
        \centering
\includegraphics[width=0.99\textwidth]{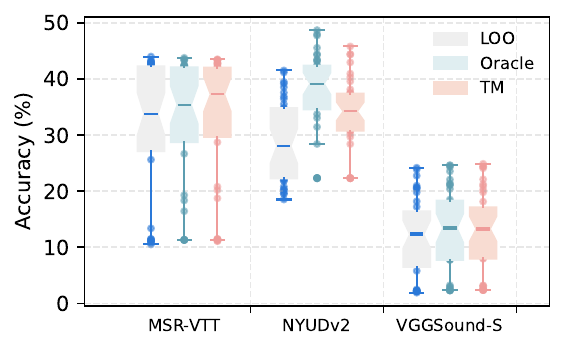}
        \vspace{-7mm}
        \caption{Averaged performance comparison for different model selection methods across different benchmarks (Weak MG). }
        \label{fig:weakMG_box}
    \end{minipage}
    \hfill
    \begin{minipage}[t]{0.64\linewidth}
        \centering
\includegraphics[width=0.99\textwidth]{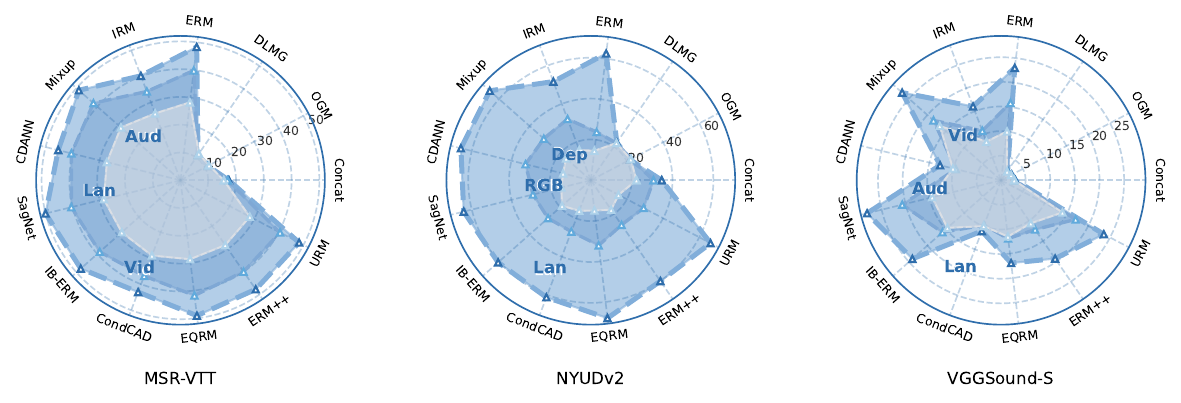}
        \vspace{-3mm}
        \caption{Weak MG Performance comparison for specific testing modalities across different algorithms.
        }
        \label{fig:weakMG_radar}
    \end{minipage}
    \vspace{-4mm}
\end{figure*}

\subsection{Implementation Details} 
We describe the implementation of modal encoding, testing perceptor implementation for Strong MG, and the architecture of the learner $h$. 

\noindent\textbf{$\bullet$ Modal encoding.}
We encode the text modality using unsupervised tokenization, while treating other modalities, such as audio and depth, as visual inputs, following recent modality-binding models~\cite{girdhar2023imagebind,zhu2024languagebind,lyu2024unibind}.
For text, we typically use a subword segmentation algorithm, such as SentencePiece~\cite{sennrich2015neural, kudo2018subword}, to convert the text into token indices. RGB images are resized and cropped to 224$\times$224 pixels. 
Following~\cite{gong2021ast} and~\cite{girdhar2022omnivore}, we encode 2-second audio sampled at 16kHz into spectrograms, which are 2D signals that can be processed as images. Depth data is converted into disparity maps for scale invariance. 
For video, we sample two frames per second and apply spatial cropping to generate sequences of 224$\times$224 images.  
Each image is then divided into 16 patches, as per ViT model~\cite{dosovitskiy2020vit}.

\noindent\textbf{$\bullet$ Testing perceptors in strong modality generalization.} 
We carefully select perceptors for the problem setting of strong modality generalization, prioritizing pre-trained models that are widely recognized within the community to facilitate training. 
Specifically, we use T5~\cite{raffel2020t5} for the text modality\footnote{\href{https://huggingface.co/google-t5/t5-small}{https://huggingface.co/google-t5/t5-small}}, and ViT~\cite{dosovitskiy2020vit} for the other modalities\footnote{\href{https://huggingface.co/google/vit-base-patch16-224}{https://huggingface.co/google/vit-base-patch16-224}}, as mentioned earlier, \textit{i.e.}, treating them as images. Notably, we incorporate a linear module to ensure a consistent embedding space, with 1024 as the default dimension. 
To adapt these models for our strong modality generalization evaluation, we retain the pre-trained weights and apply LoRA~\cite{hu2021lora} with contrastive objectives. 
Note that we use these self-supervised methods, aiming to prevent any task-specific information leakage. 
We randomly mask the samples to generate two views, which are then contrasted with other samples within the same batch. 
Here we set the batch size to 128 for the text modality and 64 for the other modalities.

\noindent\textbf{$\bullet$ Architecture of the learner $h$.}
Upon pre-trained modal perceptors (\textit{i.e.}, modality-binding models or solely trained perceptors for testing modalities), we construct a simple learner $h$ for follow-up tasks. 
Specifically, we adopt a 4-layer MLP to learn task-relevant features with ReLU as an activation function, and a linear model as a classifier to transform features to class distributions. 

We adapt various algorithms while maintaining consistency in their model architecture and optimization strategies. 
The model selection methods are implemented following \textsc{DomainBed}.
To ensure a thorough comparison, we perform an extensive hyperparameter search, randomly selecting hyperparameters and random seeds. For each algorithm, we conduct 9 trials over hyperparameter distributions (see Table~\ref{tab:hyperparameters}) with 3 different random seeds, considering parameters such as learning rate, weight decay, batch size, and other specialized hyperparameters. 

\section{Results \& Analysis}
We elaborate on the results and corresponding analysis for two types of modality generalization in this section.

\subsection{Results of Weak Modality Generalization}

The empirical results under the setting of weak modality generalization are shown in Figures~\ref{fig:weakMG_bar},~\ref{fig:weakMG_box}, and~\ref{fig:weakMG_radar}. Here we report the testing results for each dataset across three modal selection methods. 
From these results, we derive the following findings.

\noindent\textbf{$\bullet$ Performance gap between MML and DG algorithms.}
As demonstrated in Figure~\ref{fig:weakMG_bar}, the comparison between MML and DG shows that DG outperforms MML for weak modality generalization. MML typically focuses on joint embedding across modalities. This makes MML effective at leveraging the complementary information within the modalities, potentially achieving better performance in in-modal tasks. However, in the case of weak modality generalization where the model encounters unseen modalities during testing, MML struggles to generalize effectively. In contrast, DG is specifically designed to improve generalization across different domains, and its performance remains more robust when tested on unseen modalities. As a result, DG is better suited for weak modality generalization compared to MML.

\begin{figure*}
    \centering
    \includegraphics[width=\linewidth]{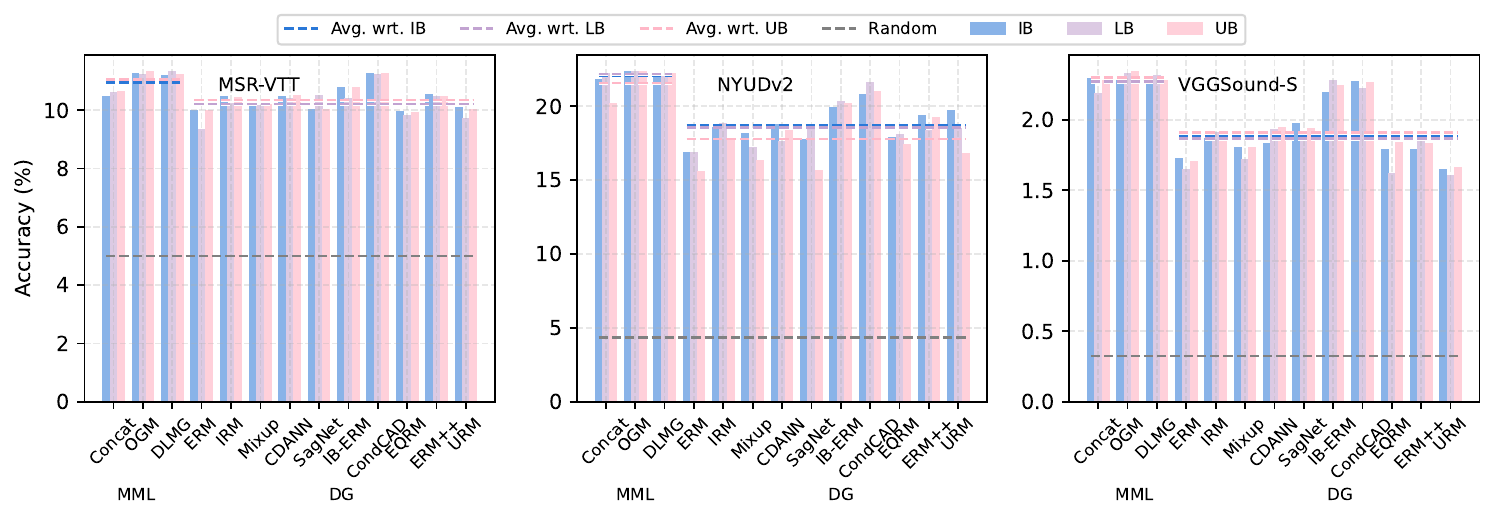}
    \vspace{-8mm}
    \caption{Strong MG performance comparison for diverse algorithms with three perceptors. 
    The dashed line indicates the averaged performance for different perceptors \textit{w.r.t.} two categories of algorithms (MML and DG). The gray line represents the performance of a random prediction. 
    The bar indicates the average performance across different testing modalities.}
    \vspace{-3mm}
    \label{fig:strongMG_bar}
\end{figure*}

\begin{figure*}[t]
    \centering
    \begin{minipage}[t]{0.34\linewidth}
        \centering
\includegraphics[width=0.99\textwidth]{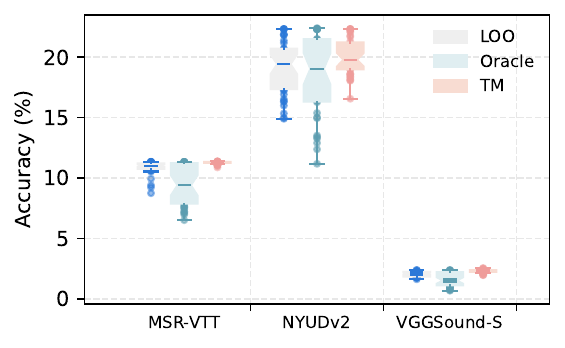}
\vspace{-7mm}
        \caption{Averaged performance comparison for different model selection methods across different benchmarks (Strong MG). }
        \label{fig:strongMG_box}
    \end{minipage}
    \hfill
    \begin{minipage}[t]{0.64\linewidth}
        \centering
\includegraphics[width=0.99\textwidth]{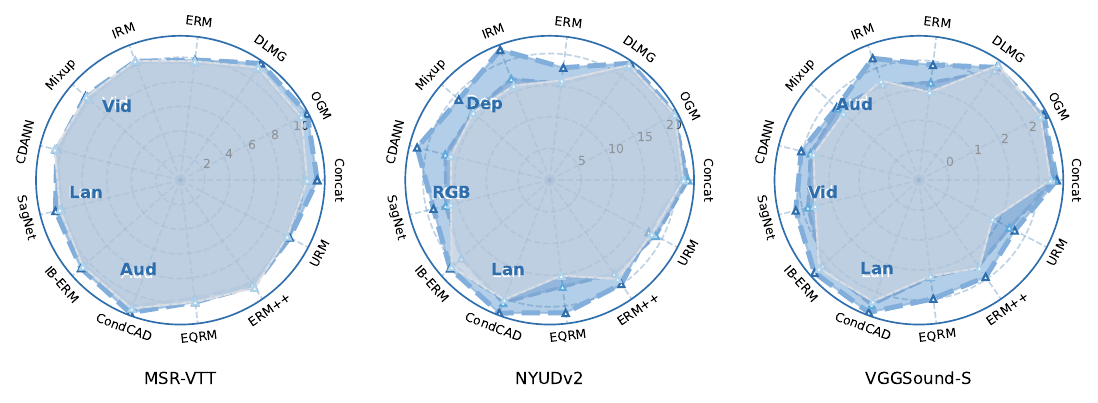}
\vspace{-3mm}
        \caption{Strong MG Performance comparison for specific testing modalities across different algorithms. }
        \label{fig:strongMG_radar}
    \end{minipage}
    
\end{figure*}

\noindent\textbf{$\bullet$ Better modality binding $\neq$ better generalization.}
Better modality binding does not necessarily translate to improved performance in modality generalization. 
For example, LanguageBind, despite achieving superior performance in modality binding, underperforms in most cases compared to the other two modality-binding methods under the Weak MG setting. 
This highlights the inherent limitations of learning a joint embedding space for modality generalization, as discussed in Section~\ref{sec:limitation-joint-embedding}.
Furthermore, this distinction manifests the core difference in learning objectives between modality binding and generalization.
Modality binding focuses on identifying the shared information across existing modalities or around specific modalities (\textit{e.g.}, image or language). 
In contrast, modality generalization seeks to capture invariants across all modalities, including novel ones. 
These differing objectives lead to inconsistencies.
This finding calls for further exploration and development of methods specifically tailored for modality generalization.

\noindent\textbf{$\bullet$ The importance of model selection.}
The performance varies significantly across different model selection methods, as shown in Figure~\ref{fig:weakMG_box}.
This is consistent with findings from \textsc{DomainBed}~\cite{gulrajani2021search}, which demonstrate that validation assumptions influence model selection criteria and ultimately determine modality generalization performance during testing. 
Among the three model selection methods, the testing-modality validation set (oracle) approach outperforms the others overall. 
By selecting the model that achieves the best performance on the validation set derived from the test modality, this method thereby optimizes the model to generalize across unseen modalities. 
These results emphasize that precise model selection is critical to the success of the experiment.

\noindent\textbf{$\bullet$ Modality differences for generalization.}
We demonstrate the performance across different testing modalities, as depicted in Figure~\ref{fig:weakMG_radar}. A clear prioritization of modalities is observed across different algorithms. For instance, generalization to ``Lan'' is notably easier on both NYUDv2 and VGGSound-S. In the Weak MG setting, existing perceptors are incorporated for their joint embedding spaces, which are typically designed for semantic purposes (\textit{e.g.}, CLIP~\cite{radford2021learning} leverages natural language to reference learned visual concepts, enhancing the development of recent modality-binding methods~\cite{girdhar2023imagebind, zhu2024languagebind, lyu2024unibind}).

\subsection{Results of Strong Modality Generalization}
We present the evaluation results under the strong modality generalization setting in Figures~\ref{fig:strongMG_bar},~\ref{fig:strongMG_box}, and~\ref{fig:strongMG_radar}.
Unlike the weak modality generalization setting, the testing perceptors are exclusively trained on data from novel modalities by self-supervised learning. From these results, we derive the following observations.

\noindent\textbf{$\bullet$ Narrowed performance gap between MML and DG.} 
In the strong modality generalization setting, the performance gap between MML and DG methods diminishes, with MML slightly outperforming DG, especially on the NYUDv2 dataset. 
This contrasts with the previous weak setting, where DG methods hold a significant advantage over MML, as shown in Figure~\ref{fig:strongMG_bar}. 
Notably, MML demonstrates more robust performance across various hyperparameter searches and model selection strategies. 
This can be attributed to its learning objective that focuses on exploiting the complementarity among multi-modalities, not aligning well with the modality generalization goal. As a comparison,
DG focuses on domain-invariant features (\textit{i.e.}, modal-invariant in our case), which are more sensitive when testing modalities are completely isolated.

\noindent\textbf{$\bullet$ Model selection remains crucial.}
The performance in the strong modality generalization setting still varies with different model selection methods, highlighting the importance of model selection in Figure~\ref{fig:strongMG_box}. Without a validation set that mirrors the test modality, hyperparameter tuning and performance benchmarking become particularly challenging. 
Among the three methods, the training-modality validation set leads to more stable performance across algorithms compared to the other two model selections. The phenomenon is different from that in the case of weak modality generalization. 
This inspires researchers to design task-specific model selection methods or more general algorithms that can consistently perform well across multiple modality generalization approaches.

\noindent\textbf{$\bullet$ Limited utility of modality binding.}
Modality binding methods improve the generalization in weak modality generalization, while exhibiting limited utility in a more challenging task, strong modality generalization.   
The alignment between the training and testing perceptors helps the model preserve the relationship between modalities, leading to better consistency and accuracy when processing novel modalities in weak modality generalization testing. 
However, their utility diminishes in strong modality generalization, where the testing modality's perceptor is isolated and does not share alignment with training modalities. 
This is due to their reliance on a joint embedding space, which struggles to generalize to unseen modalities with isolated perceptors. 
These findings suggest a need for alternative strategies that move beyond binding-centric paradigms to effectively handle Strong MG scenarios.

\noindent\textbf{$\bullet$ Narrowed generalization gap between modalities.}
In the Strong MG setting,the challenge of generalizing to the testing modality exhibits a more monotonic trend, as illustrated in Figure~\ref{fig:strongMG_radar}. The utility of modality binding, which refers to the model's capacity to associate and leverage shared features across different modalities, is notably limited in this context. 
This reduction in modality binding diminishes the distinction of ``easily generalizable'' modalities.
However, performance remains higher when ``Lan'' is the testing modality, particularly on NYUDv2 and VGGSound-S. Despite complete isolation of testing perceptors, connections between modalities can still be identified, which enhances generalization and substantiates our motivation to embrace Strong MG.

\noindent\textbf{$\bullet$ Potential of strong modality generalization.}
While the overall performance in the Strong MG setting is lower than that observed in the Weak MG setting, as evidenced by the comparative analysis in Figure~\ref{fig:strongMG_bar}), all methods achieve better-than-random predictions, indicating the ability to capture some cross-modal invariants. 
However, the performance gap between weak and strong modality generalization emphasizes the complexity of the strong modality generalization task, underlining the need for models that generalize without tightly bound perceptors or shared embedding spaces, such as those enhanced by self-supervised perceptors for modality generalization.

\section{Discussions and Outlook}
We provide several discussions to help further interpret the proposed problem setting and motivate future work.

\subsection{Modality Generalization Beyond Domain Generalization} Modality generalization (MG) extends beyond the traditional scope of domain generalization by focusing on enabling models to generalize across entirely unseen modalities rather than unseen domains. While domain generalization assumes similar underlying features across domains, MG introduces additional complexity due to the vast heterogeneity between modalities, such as text, audio, and visual data. Future research should emphasize developing strategies to learn and transfer invariant features across such diverse modalities. This includes moving beyond shared embedding spaces to identify deeper, task-agnostic representations that are transferable across unseen modalities. To address these challenges, researchers could explore the following aspects: (1) \textit{cross-modal invariant learning}: develop mechanisms to extract invariants that are robust across modalities without assuming tight alignment or correspondence; (2) \textit{hybrid generalization frameworks}: combine domain-specific techniques (\textit{e.g.}, meta-learning) with modality-aware methods to enhance adaptability and robustness; (3) \textit{realistic scenario validation}: test models under resource-constrained or privacy-sensitive scenarios where data availability is limited for certain modalities.

\subsection{Specific Algorithm Design} Designing algorithms tailored to modality generalization is critical for handling the limitations of existing multi-modal learning frameworks. First, leveraging and advancing self-supervised learning to create perceptors for unseen modalities can enable effective feature extraction even without prior alignments. Second, exploring dynamic modality binding mechanisms that adapt to the availability of training modalities and the characteristics of unseen testing modalities can enhance model flexibility. Finally, developing task-aware modality generalization algorithms that integrate task-specific priors can improve generalization for downstream applications.

\subsection{Construction of Related Datasets} The lack of comprehensive datasets for MG research remains a major challenge. Future datasets should include diverse modalities, such as text, audio, video, depth, and bio-signals, to better simulate the complexity of real-world scenarios. These datasets should be designed to reflect MG-specific challenges, intentionally excluding certain modalities during training while ensuring their presence in the testing set to create both weak and strong MG scenarios. To facilitate benchmarking for large-scale models, datasets should also vary in complexity, modality richness, and data volume, addressing scalability concerns. Additionally, real-world constraints such as privacy and resource limitations should be considered, incorporating sensitive modalities like medical imaging or encrypted communications to emphasize the practicality of MG use cases.

\subsection{Model Selection Development} Model selection is important in modality generalization. Without effective model selection, fair benchmarking and reliable performance are difficult to achieve. Future work can focus on (1) \textit{task-specific model selection}: develop and advance techniques tailored to specific MG tasks, such as leave-one-modality-out cross-validation for evaluating generalization; (2) \textit{generalized selection criteria}: design selection strategies that rely on surrogate validation sets or meta-learning approaches to ensure robust performance across diverse MG tasks; (3) \textit{adaptive hyperparameter tuning}: explore algorithms that adaptively update hyperparameters based on observed trends during training, reducing dependence on predefined validation data.

\subsection{Limitations}

This work proposes a comprehensive benchmark for the new scenario of modality generalization. In particular, we investigate five modalities across three datasets. Additional modalities in our daily lives, such as bioelectric signals, are also promising for advancing modality generalization.
Unfortunately, we cannot curate these data due to privacy concerns and data accessibility issues. These types of data often involve sensitive personal or medical information, requiring strict compliance with ethical and legal regulations. 
More modalities like thermal imaging, LiDAR scans, and haptic feedback also hold potential significance in pushing modality generalization, especially for robotics and autonomous systems. These data are scarce due to the overhead collection and labeling costs. The lack of corresponding modality encoders further hinders us from including these modalities currently. With community development, we will continue \textsc{ModalBed}, as emphasized in Section~\ref{sec:continual_modalbed}.

\newpage
\section*{Acknowledgement}
This research/project is supported by the National Research
Foundation, Singapore under its National Large Language
Models Funding Initiative (AISG Award No: AISG-NMLP2024-002). Any opinions, findings and conclusions or recommendations expressed in this material are those of the
author(s) and do not reflect the views of National Research
Foundation, Singapore. Xiaobo
Xia is supported by MoE Key Laboratory of Brain-inspired
Intelligent Perception and Cognition, University of Science
and Technology of China (Grant No. 2421002).

\bibliographystyle{ACM-Reference-Format}
\balance
\bibliography{reference}

\newpage

\appendix

\section{Implementation Details}
\label{appendix:implementation}
In this section, we elaborate on the details of the model architecture of the learner (see Table~\ref{tab:model_h}), hyperparameter setting, and other training details. 

\begin{table}[h]
\centering
\caption{Details of the learner $h$ with the output dimension of perceptors as 1024 and the number of classes as 20, 23, or 310.}
\label{tab:model_h}
\begin{tabular}{l|cc|c} 
\toprule
\# & \multicolumn{2}{c|}{\textbf{Layer}}      & \textbf{Dimensions}     \\ 
\midrule
1  & \multirow{7}{*}{\rotatebox{90}{Featurizer}} & Linear     & $1024 \rightarrow 512$  \\
2  &                             & ReLU       & -                       \\
3  &                             & Linear     & $512 \rightarrow 256$   \\
4  &                             & ReLU       & -                       \\
5  &                             & Linear     & $256 \rightarrow 512$   \\
6  &                             & ReLU       & -                       \\
7  &                             & Linear     & $512 \rightarrow 1024$  \\ 
\midrule
8  & \multicolumn{2}{c|}{Classifier (Linear)} & $1024 \rightarrow 20/23/310$  \\
\bottomrule

\end{tabular}
\end{table}

\begin{table*}[!ht]
\centering
\caption{Details of hyperparameter setting for random search. That $\mathcal{U}$ denotes the uniform distribution.}
\label{tab:hyperparameters}
\resizebox{0.8\textwidth}{!}{
\begin{tabular}{ll|lll} 
\toprule
\textbf{Type}        & \textbf{Algorithm}       & \textbf{Hyperparameter}         & \textbf{Default Value} & \textbf{Distribution}                  \\ 
\midrule
                     &                          & batchsize                       & $32$                   & $2^\mathcal{U}(3,5.5)$                 \\ 
\midrule
\multirow{5}{*}{MML} & \multirow{4}{*}{All}     & lr                              & $0.001$                & $10^\mathcal{U}(-4.5, -2.5)$           \\
                     &                          & momentum~                       & $0.9$                  & $\mathcal{U}(0.85, 0.95)$              \\
                     &                          & weight decay                    & $0.0001$               & $10^\mathcal{U}(-3.5,-4.5)$            \\
                     &                          & patience                        & $70$                   & $\mathcal{U}(60, 80)$                  \\ 
\cmidrule{2-5}
                     & OGM                      & alpha                           & $0.1$                  & $\mathcal{U}(0.1,0.3)$                 \\ 
\midrule
\multirow{20}{*}{DG} & \multirow{2}{*}{All}     & lr                              & $0.00005$              & $10^\mathcal{U}(-5,-3.5)$              \\
                     &                          & weight decay                    & $0$                    & $0$                                    \\ 
\cmidrule{2-5}
                     & \multirow{2}{*}{IRM}     & lambda                          & $100$                  & $10^\mathcal{U}(-1, 5)$                \\
                     &                          & iterations of penalty annealing & $500$                  & $10^\mathcal{U}(0,4)$                  \\ 
\cmidrule{2-5}
                     & Mixup                    & alpha                           & $0.2$                  & $10^\mathcal{U}(-1,1)$                 \\ 
\cmidrule{2-5}
                     & \multirow{5}{*}{CDANN}   & lambda                          & $1.0$                  & $10^\mathcal{U}(-2,-2)$                \\
                     &                          & discriminator weight decay      & $0$                    & $10^\mathcal{U}(-6,-2)$                \\
                     &                          & discriminator steps             & $1$                    & $2^\mathcal{U}(0,3)$                   \\
                     &                          & gradient penalty                & $0$                    & $10^\mathcal{U}(-2,1)$                 \\
                     &                          & adam beta1                      & $0.5$                  & $\{0,0.5\}$                            \\ 
\cmidrule{2-5}
                     & SagNet                   & weight of adversarial loss      & $0.1$                  & $10^\mathcal{U}(-2,1)$                 \\ 
\cmidrule{2-5}
                     & \multirow{2}{*}{IB\_ERM} & lambda                          & $100$                  & $10^\mathcal{U}(-1, 5)$                \\
                     &                          & iterations of penalty annealing & $500$                  & $10^\mathcal{U}(0,4)$                  \\ 
\cmidrule{2-5}
                     & \multirow{2}{*}{CondCAD} & lambda                          & $0.1$                  & $\{0.0001, 0.001,0.01,0.1,1,10,100\}$  \\
                     &                          & temperature                     & $0.1$                  & $\{0.05, 0.1\}$                        \\ 
\cmidrule{2-5}
                     & \multirow{3}{*}{EQRM}    & lr                              & $0.000001$             & $10^\mathcal{U}(-7, -5)$               \\
                     &                          & quantile                        & $0.75$                 & $\mathcal{U}(0.5, 0.99)$               \\
                     &                          & iterations of burn-in           & $2500$                 & $10^\mathcal{U}(2.5, 3.5)$             \\ 
\cmidrule{2-5}
                     & ERM++                    & lr                              & $0.00005$              & $10^\mathcal{U}(-5, -3.5)$             \\ 
\cmidrule{2-5}
                     & URM                      & lambda                          & $0.1$                  & $\mathcal{U}(0, 0.2)$                  \\
\bottomrule
\end{tabular}
}
\end{table*}

\noindent\textbf{Hyperparameters.}
Following \textsc{DomainBed}~\cite{gulrajani2021search}, we list the hyperparameters, their default values, and distributions for our hyperparameter sweep, while involving multimodal learning methods, in Table~\ref{tab:hyperparameters}.

\noindent\textbf{Other training details.} 
To optimize the learner $h$, we use Adam~\cite{kingma2014adam} for domain generalization algorithms and SGD~\cite{robbins1951stochastic, lecun1998gradient} for multimodal learning methods, following their official implementations. 
Moreover, we set the training steps to 5,000 for MSR-VTT and the sampled VGGSound-S dataset, and 10,000 for the larger NYUDv2 dataset.

\section{Details of Selected Algorithms}
\label{appendix:algorithms}
In this section, we detail the selected algorithms in multimodal learning and domain generalization. 
\begin{itemize}[leftmargin=*]
    \item \textbf{Concat}: Feature Concatenation integrates features from multiple modalities by combining them into a unified representation, enabling joint processing for enhanced task performance.
    \item \textbf{OGM}: On-the-fly Gradient Modulation \cite{peng2022balanced} dynamically adjusts gradients during training to balance contributions from diverse modalities, improving multi-modal learning stability.
    \item \textbf{DLMG}: Dynamically Learning Modality Gap \cite{yangfacilitating} adaptively learns to bridge modality-specific differences, facilitating effective integration of heterogeneous data for robust task outcomes.
    \item \textbf{ERM}: Empirical Risk Minimization \cite{vapnik1998statistical} minimizes average loss across training domains, serving as a baseline for domain generalization tasks.
    \item \textbf{IRM}: Invariant Risk Minimization \cite{arjovsky2019invariant} seeks invariant predictors across domains, ensuring robust generalization to unseen domains.
    \item \textbf{Mixup}: Inter-domain Mixup \cite{yan2020improve} blends data across domains via interpolation, enhancing model robustness and generalization.
    \item \textbf{CDANN}: Class-conditional DANN \cite{li2018deep} extends adversarial domain adaptation by aligning class-conditional distributions, improving cross-domain performance.
    \item \textbf{SagNet}: Style Agnostic Networks \cite{nam2021reducing} separate style and content, reducing domain-specific biases for better generalization.
    \item \textbf{IB$_\text{ERM}$}: Information Bottleneck \cite{ahuja2021invariance} applies the bottleneck principle to ERM, minimizing irrelevant information for domain-invariant learning.
    \item \textbf{CondCAD}: Conditional Contrastive Adversarial Domain \cite{ruan2021optimal} leverages contrastive learning and adversarial methods to align conditional distributions across domains.
    \item \textbf{EQRM}: Empirical Quantile Risk Minimization \cite{eastwood2022eqrm} optimizes risk across quantiles, enhancing robustness to domain shifts.
    \item \textbf{ERM++}: Improved Empirical Risk Minimization \cite{teterwak2023erm++} refines ERM with advanced techniques for better cross-domain generalization.
    \item \textbf{URM}: Uniform Risk Minimization \cite{krishnamachariuniformly} targets uniform risk across domains, ensuring consistent performance in varied settings.
\end{itemize}

\section{Details of Model Selection Methods}
\label{appendix:model_selection}
In this section, we detail the three model selection methods for evaluating modality generalization performance. 

\noindent\textbf{Training-modality validation set.}
By assuming that the training and testing samples follow the same distributions, we directly construct the validation set derived from the training dataset across each modality.
To this end, we choose the model that maximizes the accuracy within this validation set.

\noindent\textbf{Leave-one-modality-out cross-validation.}
We iteratively leave one modality out of the training set to serve as the validation modality, while the remaining modalities are used to train the model. For each training modality, a separate model is trained and evaluated on its corresponding held-out modality. The performance metrics (\textit{i.e.}, accuracy) are averaged across all held-out modalities, and the model with the highest average accuracy is selected. Finally, this chosen model is retrained using all the training modalities to maximize generalization performance on unseen test modalities. This approach assumes that training and test modalities originate from a shared \emph{meta-distribution} and aims to optimize expected performance across this \emph{meta-distribution}.

\noindent\textbf{Test-modality validation set (oracle).}
We select models based on their accuracy on a validation set derived from the test domain distribution. It permits up to multiple queries during hyperparameter tuning with each query corresponding to one hyperparameter configuration. 
All models are trained for a fixed number of steps without early stopping, and only the final checkpoint is evaluated. While this method can guide model selection, it assumes access to the test modalities and limits the number of hyperparameter combinations~\cite{gulrajani2021search}.

\section{Detailed Experimental Results}
\label{appendix:results}
Here we provide experimental results in detail, especially including the results over different modalities. 
Specifically, we include the results for three multimodal learning methods and 10 domain generalization methods, which are equipped with three different modality bindings as perceptors.
Three model selection methods (detailed in Appendix~\ref{appendix:model_selection}), are used to evaluate the above methods across three datasets, involving MST-VTT, NYUDv2, and VGGSound-S. 
Furthermore, different datasets consist of different modalities, emphasizing the diversity of our benchmarking. 
For a comprehensive demonstration, the results of two modality generalization settings are shown using 6 tables according to different model selection methods, with Tables~\ref{tab:weak_mg1}, \ref{tab:weak_mg2}, and \ref{tab:weak_mg3} are for Weak MG, and Tables~\ref{tab:strong_mg1}, \ref{tab:strong_mg2}, and \ref{tab:strong_mg3} are for Strong MG. These experimental results also align with the illustrations in Figures~\ref{fig:weakMG_bar}, ~\ref{fig:weakMG_box}, ~\ref{fig:weakMG_radar}, ~\ref{fig:strongMG_bar}, ~\ref{fig:strongMG_box},  and~\ref{fig:strongMG_radar}.

 \begin{table*}[!h]
\centering
\caption{Mean and standard deviation of classification performance comparison under the Weak MG setting \textit{w.r.t.} training-modality validation set as model selection.}
\label{tab:weak_mg1}
\resizebox{\textwidth}{!}{ 
\setlength{\tabcolsep}{2mm}
\begin{tabular}{ccc|llll|llll|llll}
\toprule
\multicolumn{15}{c}{Weak MG, Model selection: training-modality validation set}\\
\midrule
\multirow{2}{*}{\textbf{Perceptor}}& \multicolumn{2}{c|}{\multirow{2}{*}{\textbf{Method}}} & \multicolumn{4}{c|}{\textbf{ MSR-VTT }}  & \multicolumn{4}{c|}{\textbf{ NYUDv2 }}  & \multicolumn{4}{c}{\textbf{ VGGSound-S }}  \\
\cmidrule{4-15}
& & & \multicolumn{1}{c}{\textbf{Vid}} & \multicolumn{1}{c}{\textbf{Aud}} & \multicolumn{1}{c}{\textbf{Lan}} & \multicolumn{1}{c|}{\textbf{Avg}} & \multicolumn{1}{c}{\textbf{Dep}} & \multicolumn{1}{c}{\textbf{RGB}} & \multicolumn{1}{c}{\textbf{Lan}} & \multicolumn{1}{c|}{\textbf{Avg}} & \multicolumn{1}{c}{\textbf{Vid}} & \multicolumn{1}{c}{\textbf{Aud}} & \multicolumn{1}{c}{\textbf{Lan}} & \multicolumn{1}{c}{\textbf{Avg}} \\
\midrule
\multicolumn{1}{c}{\multirow{13}{*}{\rotatebox{90}{ImageBind}}} & \multicolumn{1}{c}{\multirow{3}{*}{MML}} & \multicolumn{1}{l|}{Concat} &\multicolumn{1}{c}{$\text{23.3}_{\pm\text{3.0}}$} & \multicolumn{1}{c}{$\text{16.9}_{\pm\text{1.0}}$} & \multicolumn{1}{c}{$\text{20.4}_{\pm\text{1.4}}$} & \multicolumn{1}{c|}{\text{20.2}} & \multicolumn{1}{c}{$\text{23.7}_{\pm\text{1.3}}$} & \multicolumn{1}{c}{$\text{32.6}_{\pm\text{4.4}}$} & \multicolumn{1}{c}{$\text{32.8}_{\pm\text{4.2}}$} & \multicolumn{1}{c|}{\text{29.7}} & \multicolumn{1}{c}{$\text{3.5}_{\pm\text{0.6}}$} & \multicolumn{1}{c}{$\text{2.9}_{\pm\text{0.4}}$} & \multicolumn{1}{c}{$\text{3.1}_{\pm\text{0.6}}$} & \multicolumn{1}{c}{\text{3.2}} \\
\end{tabular}
}
\end{table*}
\begin{table*}[!h]
\centering
\caption{Mean and standard deviation of classification performance comparison under the Weak MG setting \textit{w.r.t.} leave-one-modality-out cross-validation as model selection.}
\label{tab:weak_mg2}
\resizebox{\textwidth}{!}{ 
\setlength{\tabcolsep}{2mm}
\begin{tabular}{ccc|llll|llll|llll}
\toprule
\multicolumn{15}{c}{Weak MG, Model selection: leave-one-modality-out cross-validation}\\
\midrule
\multirow{2}{*}{\textbf{Perceptor}}& \multicolumn{2}{c|}{\multirow{2}{*}{\textbf{Method}}} & \multicolumn{4}{c|}{\textbf{ MSR-VTT }}  & \multicolumn{4}{c|}{\textbf{ NYUDv2 }}  & \multicolumn{4}{c}{\textbf{ VGGSound-S }}  \\
\cmidrule{4-15}
& & & \multicolumn{1}{c}{\textbf{Vid}} & \multicolumn{1}{c}{\textbf{Aud}} & \multicolumn{1}{c}{\textbf{Lan}} & \multicolumn{1}{c|}{\textbf{Avg}} & \multicolumn{1}{c}{\textbf{Dep}} & \multicolumn{1}{c}{\textbf{RGB}} & \multicolumn{1}{c}{\textbf{Lan}} & \multicolumn{1}{c|}{\textbf{Avg}} & \multicolumn{1}{c}{\textbf{Vid}} & \multicolumn{1}{c}{\textbf{Aud}} & \multicolumn{1}{c}{\textbf{Lan}} & \multicolumn{1}{c}{\textbf{Avg}} \\
\midrule
\multicolumn{1}{c}{\multirow{13}{*}{\rotatebox{90}{ImageBind}}} & \multicolumn{1}{c}{\multirow{3}{*}{MML}} & \multicolumn{1}{l|}{Concat} &\multicolumn{1}{c}{$\text{11.1}_{\pm\text{0.0}}$} & \multicolumn{1}{c}{$\text{11.1}_{\pm\text{0.6}}$} & \multicolumn{1}{c}{$\text{12.2}_{\pm\text{1.0}}$} & \multicolumn{1}{c|}{\text{11.5}} & \multicolumn{1}{c}{$\text{23.0}_{\pm\text{0.6}}$} & \multicolumn{1}{c}{$\text{25.2}_{\pm\text{2.4}}$} & \multicolumn{1}{c}{$\text{22.4}_{\pm\text{0.0}}$} & \multicolumn{1}{c|}{\text{23.5}} & \multicolumn{1}{c}{$\text{2.3}_{\pm\text{0.0}}$} & \multicolumn{1}{c}{$\text{1.9}_{\pm\text{0.2}}$} & \multicolumn{1}{c}{$\text{2.3}_{\pm\text{0.0}}$} & \multicolumn{1}{c}{\text{2.2}} \\
\end{tabular}
}
\end{table*}
\begin{table*}[!h]
\centering
\caption{Mean and standard deviation of classification performance comparison under the Weak MG setting \textit{w.r.t.} test-modality validation set (oracle) as model selection.}
\label{tab:weak_mg3}
\resizebox{\textwidth}{!}{ 
\setlength{\tabcolsep}{2mm}
\begin{tabular}{ccc|llll|llll|llll}
\toprule
\multicolumn{15}{c}{Weak MG, Model selection: test-modality validation set (oracle)}\\
\midrule
\multirow{2}{*}{\textbf{Perceptor}}& \multicolumn{2}{c|}{\multirow{2}{*}{\textbf{Method}}} & \multicolumn{4}{c|}{\textbf{ MSR-VTT }}  & \multicolumn{4}{c|}{\textbf{ NYUDv2 }}  & \multicolumn{4}{c}{\textbf{ VGGSound-S }}  \\
\cmidrule{4-15}
& & & \multicolumn{1}{c}{\textbf{Vid}} & \multicolumn{1}{c}{\textbf{Aud}} & \multicolumn{1}{c}{\textbf{Lan}} & \multicolumn{1}{c|}{\textbf{Avg}} & \multicolumn{1}{c}{\textbf{Dep}} & \multicolumn{1}{c}{\textbf{RGB}} & \multicolumn{1}{c}{\textbf{Lan}} & \multicolumn{1}{c|}{\textbf{Avg}} & \multicolumn{1}{c}{\textbf{Vid}} & \multicolumn{1}{c}{\textbf{Aud}} & \multicolumn{1}{c}{\textbf{Lan}} & \multicolumn{1}{c}{\textbf{Avg}} \\
\midrule
\multicolumn{1}{c}{\multirow{13}{*}{\rotatebox{90}{ImageBind}}} & \multicolumn{1}{c}{\multirow{3}{*}{MML}} & \multicolumn{1}{l|}{Concat} &\multicolumn{1}{c}{$\text{19.1}_{\pm\text{4.0}}$} & \multicolumn{1}{c}{$\text{15.6}_{\pm\text{1.8}}$} & \multicolumn{1}{c}{$\text{20.4}_{\pm\text{1.4}}$} & \multicolumn{1}{c|}{\text{18.4}} & \multicolumn{1}{c}{$\text{22.8}_{\pm\text{0.4}}$} & \multicolumn{1}{c}{$\text{27.3}_{\pm\text{2.6}}$} & \multicolumn{1}{c}{$\text{44.3}_{\pm\text{8.9}}$} & \multicolumn{1}{c|}{\text{31.5}} & \multicolumn{1}{c}{$\text{3.0}_{\pm\text{0.5}}$} & \multicolumn{1}{c}{$\text{2.9}_{\pm\text{0.3}}$} & \multicolumn{1}{c}{$\text{3.0}_{\pm\text{0.6}}$} & \multicolumn{1}{c}{\text{3.0}} \\
\end{tabular}
}
\end{table*}

\begin{table*}[!h]
\centering
\caption{Mean and standard deviation of classification performance comparison under the Strong MG setting \textit{w.r.t.} training-modality validation set as model selection.}
\label{tab:strong_mg1}
\resizebox{\textwidth}{!}{ 
\setlength{\tabcolsep}{2mm}
\begin{tabular}{ccc|llll|llll|llll}
\toprule
\multicolumn{15}{c}{Strong MG, Model selection: training-modality validation set}\\
\midrule
\multirow{2}{*}{\textbf{Perceptor}}& \multicolumn{2}{c|}{\multirow{2}{*}{\textbf{Method}}} & \multicolumn{4}{c|}{\textbf{ MSR-VTT }}  & \multicolumn{4}{c|}{\textbf{ NYUDv2 }}  & \multicolumn{4}{c}{\textbf{ VGGSound-S }}  \\
\cmidrule{4-15}
& & & \multicolumn{1}{c}{\textbf{Vid}} & \multicolumn{1}{c}{\textbf{Aud}} & \multicolumn{1}{c}{\textbf{Lan}} & \multicolumn{1}{c|}{\textbf{Avg}} & \multicolumn{1}{c}{\textbf{Dep}} & \multicolumn{1}{c}{\textbf{RGB}} & \multicolumn{1}{c}{\textbf{Lan}} & \multicolumn{1}{c|}{\textbf{Avg}} & \multicolumn{1}{c}{\textbf{Vid}} & \multicolumn{1}{c}{\textbf{Aud}} & \multicolumn{1}{c}{\textbf{Lan}} & \multicolumn{1}{c}{\textbf{Avg}} \\
\midrule
\multicolumn{1}{c}{\multirow{13}{*}{\rotatebox{90}{ImageBind}}} & \multicolumn{1}{c}{\multirow{3}{*}{MML}} & \multicolumn{1}{l|}{Concat} &\multicolumn{1}{c}{$\text{11.4}_{\pm\text{0.3}}$} & \multicolumn{1}{c}{$\text{11.5}_{\pm\text{0.3}}$} & \multicolumn{1}{c}{$\text{11.0}_{\pm\text{0.0}}$} & \multicolumn{1}{c|}{\text{11.3}} & \multicolumn{1}{c}{$\text{21.9}_{\pm\text{0.2}}$} & \multicolumn{1}{c}{$\text{20.8}_{\pm\text{0.7}}$} & \multicolumn{1}{c}{$\text{22.4}_{\pm\text{0.0}}$} & \multicolumn{1}{c|}{\text{21.7}} & \multicolumn{1}{c}{$\text{2.3}_{\pm\text{0.0}}$} & \multicolumn{1}{c}{$\text{2.1}_{\pm\text{0.1}}$} & \multicolumn{1}{c}{$\text{2.3}_{\pm\text{0.0}}$} & \multicolumn{1}{c}{\text{2.3}} \\
\end{tabular}
}
\end{table*}
\begin{table*}[!h]
\centering
\caption{Mean and standard deviation of classification performance comparison under the Strong MG setting \textit{w.r.t.} leave-one-modality-out cross-validation as model selection.}
\label{tab:strong_mg2}
\resizebox{\textwidth}{!}{ 
\setlength{\tabcolsep}{2mm}
\begin{tabular}{ccc|llll|llll|llll}
\toprule
\multicolumn{15}{c}{Strong MG, Model selection: leave-one-modality-out cross-validation}\\
\midrule
\multirow{2}{*}{\textbf{Perceptor}}& \multicolumn{2}{c|}{\multirow{2}{*}{\textbf{Method}}} & \multicolumn{4}{c|}{\textbf{ MSR-VTT }}  & \multicolumn{4}{c|}{\textbf{ NYUDv2 }}  & \multicolumn{4}{c}{\textbf{ VGGSound-S }}  \\
\cmidrule{4-15}
& & & \multicolumn{1}{c}{\textbf{Vid}} & \multicolumn{1}{c}{\textbf{Aud}} & \multicolumn{1}{c}{\textbf{Lan}} & \multicolumn{1}{c|}{\textbf{Avg}} & \multicolumn{1}{c}{\textbf{Dep}} & \multicolumn{1}{c}{\textbf{RGB}} & \multicolumn{1}{c}{\textbf{Lan}} & \multicolumn{1}{c|}{\textbf{Avg}} & \multicolumn{1}{c}{\textbf{Vid}} & \multicolumn{1}{c}{\textbf{Aud}} & \multicolumn{1}{c}{\textbf{Lan}} & \multicolumn{1}{c}{\textbf{Avg}} \\
\midrule
\multicolumn{1}{c}{\multirow{13}{*}{\rotatebox{90}{ImageBind}}} & \multicolumn{1}{c}{\multirow{3}{*}{MML}} & \multicolumn{1}{l|}{Concat} &\multicolumn{1}{c}{$\text{8.8}_{\pm\text{1.7}}$} & \multicolumn{1}{c}{$\text{10.1}_{\pm\text{1.4}}$} & \multicolumn{1}{c}{$\text{7.4}_{\pm\text{1.3}}$} & \multicolumn{1}{c|}{\text{8.7}} & \multicolumn{1}{c}{$\text{21.0}_{\pm\text{1.0}}$} & \multicolumn{1}{c}{$\text{22.3}_{\pm\text{0.0}}$} & \multicolumn{1}{c}{$\text{20.3}_{\pm\text{1.7}}$} & \multicolumn{1}{c|}{\text{21.2}} & \multicolumn{1}{c}{$\text{2.2}_{\pm\text{0.1}}$} & \multicolumn{1}{c}{$\text{2.1}_{\pm\text{0.2}}$} & \multicolumn{1}{c}{$\text{2.3}_{\pm\text{0.0}}$} & \multicolumn{1}{c}{\text{2.2}} \\
\end{tabular}
}
\end{table*}
\begin{table*}[!h]
\centering
\caption{Mean and standard deviation of classification performance comparison under the Strong MG setting \textit{w.r.t.} test-modality validation set (oracle) as model selection.}
\label{tab:strong_mg3}
\resizebox{\textwidth}{!}{ 
\setlength{\tabcolsep}{2mm}
\begin{tabular}{ccc|llll|llll|llll}
\toprule
\multicolumn{15}{c}{Strong MG, Model selection: test-modality validation set (oracle)}\\
\midrule
\multirow{2}{*}{\textbf{Perceptor}}& \multicolumn{2}{c|}{\multirow{2}{*}{\textbf{Method}}} & \multicolumn{4}{c|}{\textbf{ MSR-VTT }}  & \multicolumn{4}{c|}{\textbf{ NYUDv2 }}  & \multicolumn{4}{c}{\textbf{ VGGSound-S }}  \\
\cmidrule{4-15}
& & & \multicolumn{1}{c}{\textbf{Vid}} & \multicolumn{1}{c}{\textbf{Aud}} & \multicolumn{1}{c}{\textbf{Lan}} & \multicolumn{1}{c|}{\textbf{Avg}} & \multicolumn{1}{c}{\textbf{Dep}} & \multicolumn{1}{c}{\textbf{RGB}} & \multicolumn{1}{c}{\textbf{Lan}} & \multicolumn{1}{c|}{\textbf{Avg}} & \multicolumn{1}{c}{\textbf{Vid}} & \multicolumn{1}{c}{\textbf{Aud}} & \multicolumn{1}{c}{\textbf{Lan}} & \multicolumn{1}{c}{\textbf{Avg}} \\
\midrule
\multicolumn{1}{c}{\multirow{13}{*}{\rotatebox{90}{ImageBind}}} & \multicolumn{1}{c}{\multirow{3}{*}{MML}} & \multicolumn{1}{l|}{Concat} &\multicolumn{1}{c}{$\text{11.1}_{\pm\text{0.0}}$} & \multicolumn{1}{c}{$\text{11.8}_{\pm\text{0.1}}$} & \multicolumn{1}{c}{$\text{11.0}_{\pm\text{0.0}}$} & \multicolumn{1}{c|}{\text{11.3}} & \multicolumn{1}{c}{$\text{22.3}_{\pm\text{0.0}}$} & \multicolumn{1}{c}{$\text{22.3}_{\pm\text{0.0}}$} & \multicolumn{1}{c}{$\text{22.4}_{\pm\text{0.0}}$} & \multicolumn{1}{c|}{\text{22.3}} & \multicolumn{1}{c}{$\text{2.3}_{\pm\text{0.0}}$} & \multicolumn{1}{c}{$\text{2.4}_{\pm\text{0.0}}$} & \multicolumn{1}{c}{$\text{2.3}_{\pm\text{0.0}}$} & \multicolumn{1}{c}{\text{2.4}} \\
\end{tabular}
}
\end{table*}

\end{document}